\documentclass[conference]{IEEEtran}
\IEEEoverridecommandlockouts
\usepackage{cite}
\usepackage{amsmath,amssymb,amsfonts}
\usepackage{algorithmic}
\usepackage{graphicx}
\usepackage{textcomp}
\usepackage{xcolor}
\usepackage{booktabs}
\usepackage{epsfig}
\usepackage{multicol}

\usepackage{caption}
\usepackage{subcaption}

\usepackage[breaklinks]{hyperref}
\hypersetup{
    colorlinks=true,
    citecolor=blue,
    linkcolor =	red,
    filecolor=magenta,      
    urlcolor=cyan,
}

\usepackage{multirow,tabularx,capt-of}
\def\BibTeX{{\rm B\kern-.05em{\sc i\kern-.025em b}\kern-.08em
    T\kern-.1667em\lower.7ex\hbox{E}\kern-.125emX}}
\begin{document}

\title{2D Pose Estimation based Child Action Recognition}
% \title{Comparison of RGB modality and Skeleton modality in Child Action Recognition}

%{Comparison of RGB based modality and Skeleton based modality in Child Action Recognition} <- first title
%% above is a [tentative title] 
%%change the title as needed

\author{
\IEEEauthorblockN{Sanka Mohottala, Sandun Abeygunawardana, Pradeepa Samarasinghe, Dharshana Kasthurirathna} 
\IEEEauthorblockA{\textit{Faculty of Computing, Sri Lanka Institute of Information Technology, Sri Lanka} \\
%\textit{Sri Lanka Institute of }\\
%\textit{Information Technology}\\
%Sri Lanka\\
sanka.m@sliit.lk, it19362854@my.sliit.lk, pradeepa.s@sliit.lk, dharshana.k@sliit.lk}
\\
\IEEEauthorblockN{Charith Abhayaratne} 
\IEEEauthorblockA{\textit{Department of Electronic and Electrical Engineering, University of Sheffield, United Kingdom} \\
%\textit{Electrical Engineering,}\\
%\textit{University of Sheffield}\\
%United Kingdom \\
c.abhayaratne@sheffield.ac.uk}
}

\maketitle

\begin{abstract}
We present a graph convolutional network with 2D pose estimation for the first time on child action recognition task achieving on par results with LRCN on a benchmark dataset containing unconstrained environment based videos.

%based implementation achieving on par results with LRCN model on child action recognition task using a benchmark dataset containing unconstrained environment based videos with the use of OpenPose.  

\end{abstract}


%%%%%%%%%%%%%%%%%%%%%%%%%%%%%%%%%%%%%%%%%%%%%%%%%%%%%%%%%%%%%%%%%%%%%%%%%%%%%%%%

%-------------------------------------------------------------------------
\section{Introduction}
\label{section:intro}

%Pradeepa reviewed

Human activity recognition (HAR) has been a focused research area due to its diverse applications in human-computer interaction~\cite{7803262}, surveillance~\cite{SINGH2017265} and health care~\cite{8463470}. The main goal of HAR is to identify actions performed by one or more humans in a temporal sequence of observations. The increased computational power together with the availability of public datasets have enabled the training of large networks, such as multi-stream 3D Convolutional Neural Network (CNN) architectures~\cite{8954434}, significantly boosting action recognition performance. 
%However, human activity recognition research has so far been primarily focused on adults.

Child action recognition (CAR) has important applications in safety monitoring~\cite{6628549}, development assessment~\cite{westeynAbowd}, and others. However, children are largely underrepresented in both pose estimation datasets~\cite{sciortino_estimation_2017} and HAR datasets. This can be explained by the fact that data from adult activities can be used in many more applications than data from children activities. Moreover, some of the largest annotated human activity datasets rely on user uploaded videos to public platforms~\cite{soomro2012ucf101},~\cite{DBLP:journals/corr/KayCSZHVVGBNSZ17}, in which
the presence of children is limited due to privacy concerns.

RGB modality being one of the oldest and most used method of acquiring motion information, many deep learning techniques on HAR have achieved best accuracy on RGB benchmark datasets like UCF101 and Kinetics600~\cite{soomro2012ucf101},~\cite{carreira2018short}. Among those state-of-the-art  methods, Long-Term Recurrent Convolutional Network (LRCN) architecture was selected as the best approach in this study for the HAR~\cite{donahue2015long} due to it's capability in handling  variable length input  and  learning complex video sequences.

%Human action recognition (HAR) using RGB videos is a challenging task. The modern evaluation of Convolutional Neural Network (CNN) and Long-Short Term Memory Network (LSTM) models has reduced the complexity of HAR using RGB videos.Recent studies focused on HAR using RGB videos and conducted with different methods and achieved best accuracy on benchmarked datasets like UCF101 and Kinetics600~\cite{soomro2012ucf101}, ~\cite{carreira2018short}.

%The process of variable length input sequences and variable length outputs, as well as the generation of full-length sequence descriptions, extends beyond the traditional one-versus-all prediction tasks. Long-Term Recurrent Convolutional Network (LRCN) instantiates for specific video activity recognition, image caption generation and video description tasks. Among those state-of-the-art methods, LRCN architecture was selected as the best approach in this study for the HAR ~\cite{donahue2015long}. LRCN has used a video classification to determine the actions from the RGB videos.

%%we need to say by 'RGB modality' we mean unchanged RGB videos as the input.
%% reason 1: there is a comparison between skeleton modality and OpticalFlow+Gray
%% reason 2: feature extraction on RGB+Deep Learning always performs better even on NTU RGB+D dataset by a large margin

%first contribution
% why STGCN
%2nd contribution

%here include the RGB modality information (i.e., LRCN)
While RGB modality has been the prominent approach for HAR, different non-RGB approachs have also come into play in the past decade with the revival of deep learning. Veeriah et al.~\cite{veeriah2015differential} introduced a modified LSTM approach where HOG3D was used to extract features from the image and they were used as the input to the model. Fernando et al.~\cite{Fernando2015ModelingVE} introduced novel feature extraction methods and a ranking machine approach to learn a hierarchical representation of actions from video frames. These approaches have been used with HAR datasets such as Kinetics-400~\cite{kay2017kinetics} and have achieved comparable results~\cite{shi2019two, yan2018spatial}. 

Human pose estimation through 2D keypoint estimation is another feature extraction approach that has been popularised recently due to emergence of libraries such as OpenPose~\cite{cao2017realtime}, and BlazePose~\cite{bazarevsky2020blazepose} where the resultant features resemble the human skeleton, often referred as the skeleton modality. While skeleton modality approaches with LSTM~\cite{shahroudy2016ntu} and temporal convolution models (TCN)~\cite{kim2017tcn} have outperformed other non-RGB approaches, graph convolution (GCN) approaches such as ST-GCN~\cite{yan2018spatial}, 2s-AGCN~\cite{shi2019two} have surpassed both these approaches. ST-GCN being the first GCN implementation for HAR, in this research, we use it for skeleton modality implementations. OpenPose was used for pose estimation since~\cite{sciortino_estimation_2017} shows that in child pose estimation, OpenPose performs better than other methods even when truncations and occlusions are present.

Most of the CAR research done with skeleton modality are based on RGB+D datasets but there are few using RGB datasets and pose estimation techniques. An LSTM based model for stereotypical action recognition of children for ASD detection is introduced in~\cite{s21020411}, and~\cite{suzuki2019enhancement, suzuki2020enhancement} recognized gross motor actions of 4-5 years old children with OpenPose as the pose estimator. %The dataset has been created in a constrained environment but the 
Authors in~\cite{s21020411} have introduced methods to overcome truncation and occlusion issues %and show an increase of $\approx$10\% in accuracy 
resulting in an overall 90\% accuracy. % as well and achieves accuracy around 90\%.%they have removed the frame if the missing joints are important joints (ex: shoulders,neck etc.) and if not then they have used the previous joint’s position as the new one.
 For tracking, they have utilized the distance of corresponding skeleton joints in frame sequence. %In~\cite{suzuki2019enhancement, suzuki2020enhancement} , the authors have used OpenPose on RGB videos for gross motor action recognition of 4-5 years old children and the dataset used is created in a constrained environment.
 A CNN approach has been used in~\cite{suzuki2019enhancement} achieving 82\% accuracy, where tracking was done using a particle filter algorithm and skeleton standardization was applied such that the output looks as if the camera angle and skeleton size remain the same. %They have achieved 82\% accuracy. 
Improving this further the authors have introduced a model similar to TCN in~\cite{suzuki2020enhancement}, where an improved standardization and a novel data augmentation were utilized to achieve $\approx$99\% accuracy. All these CAR implementations in skeleton modality have used data captured in constrained environments, hence they work with stabilized skeleton sequences. Since our interest is in unconstrained environments, we are using annotated subsets from Kinetics-400 and Kinetics-600 datasets in this research.

Most of the datasets used in CAR research are either private datasets as in~\cite{suzuki2019enhancement,suzuki2020enhancement,s21020411} and the openly available ones such as~\cite{Vatavu2019CFBG,DongYuzhu2020kindergator} contain only the skeleton data. While there are handful of datasets available such as~\cite{Rajagopalan2013,Aloba2018kindergator} they contain only a small number of videos. With the release of our annotated dataset along with the benchmark results detailed in the results section, we provide opportunities for further research on annotated public datasets.

As evidenced through the inferior performance of C3D and Resnet50+LSTM models in~\cite{baradel2018glimpse}, the skeleton modality has outperformed RGB modality on datasets created in constrained environments~\cite{shahroudy2016ntu}. But it's performance on unconstrained environment based datasets~\cite{kay2017kinetics}  is inferior to RGB modality as shown by ST-GCN and PoseC3D models in~\cite{yan2018spatial,duan2022revisiting}. While these datasets are mostly adult based datasets,~\cite{yan2018spatial} hypothesize that this performance degradation is due to information loss of object/scene interaction and shows through a Kinetics-400 subset called Kinetics-motion that when the activities are highly motion oriented, skeleton modality achieves performance similar to RGB modality. Motivated by this insight, in this research we attempt to achieve comparable results on unconstrained skeleton modality child activity datasets.

%newly added parts -2022-7-29
With the changes done to the model architecture, OpenPose V2 (2019) has improved accuracy by 7\% compared to OpenPose V1 (2016)~\cite{cao2017realtime}. OpenPose outputs the locations of skeleton joint along with a confidence value that give a measure about the reliability of the inferred joint. We attempt to see if there is any correlation between accuracy of models and the average confidence values.     

Based on the above literature study and the gaps identified, in this paper we provide the following main contributions. 
% \begin{itemize}
% \item To the best of our knowledge this is the first comparison done on skeleton-modality and the RGB-modality  based deep learning models in child activity recognition.  
% \item We provide results on a benchmark dataset along with detailed guideline to continue further research.  
% \item we show that when pre-trained with large datasets, ST-GCN on skeleton modality achieves on par accuracy with LRCN on RGB modality.
% \item we show there is no strong correlation between confidence value and accuracy %atm not included in intro
% \item we show there is no correlation between improved key-point extraction approaches and accuracy 

% \end{itemize}
\begin{itemize}
\item To the best of our knowledge this is the first GCN based CAR implementation done on extracted 2D skeleton sequences from unconstrained environment videos  
\item We show that when pre-trained with large datasets, ST-GCN on skeleton modality achieves on par accuracy with LRCN on RGB modality.
\item We show there is no strong correlation between confidence value of skeleton sequences and the class-wise accuracy of the model. %atm not included in intro
\item We provide the annotated dataset along with detailed guideline to continue further research\footnote{Dataset and other resources: \href{https://github.com/sankadivandya/KS-KSS-Dataset}{github.com/sankadivandya/KS-KSS-Dataset}}.  
% \item we show there is no correlation between improved key-point extraction approaches and accuracy 
\end{itemize}

The rest of the paper is organized as follows. 
The implemented model architectures and datasets are discussed in Section~\ref{section:methodology}. Experimental setups used in each case and the evaluation methods are discussed in Section~\ref{section:experiments}. The performance of different approaches and comparison of models are discussed in Section~\ref{section:results,discussion} while Section~\ref{section:conclusion}
concludes with future research directions.
\\

%Child Action Recognition (CAR) was not a specifically targeted area because the HAR domain encompasses the CAR component as well in RGB video-based analysis methods. Typically, most of the studies are not focused on RGB video methods in CAR~\cite{suzuki2020enhancement},~\cite{suzuki2019enhancement}.However, CAR with RGB videos was done by the same method used in HAR with the benchmarked kinetcs600 dataset was used to conduct the CAR and achieved good accuracy compared with the Adult Action Recognition (AAR).

\section{Methodology}
\label{section:methodology}

%need justifications for using LRCN and STGCN? why not recent better models? they are not comparable. they are not trained on the same dataset in according to their papers
% argue they are breakthrough models;one include GNN and other both CNN and LSTM
%CNN is upgraded with ResNet model

\subsection{Datasets}
\label{section:Datasets}
%section name has to be improved
% dataset annotation - since it was done by our side
Due to the scarcity of public child datasets, an annotated child dataset was created using an eight class subset of Kinetics-600 dataset. Based on the appearance of the main performer of the action, each video was labeled as child or adult. Kinetics-600 dataset was chosen since the activities are already classified and time stamped and since the videos are taken from YouTube, they belong to real-world scenarios. Eight classes that were selected are detailed in Table~\ref{T:child_adult_kinetics_600}. They were selected based on the similarity to Child-Whole Body Gesture (CWBG) dataset~\cite{Vatavu2019CFBG}. Majority of selected classes are motion oriented actions since we are interested in such actions as detailed in Section~\ref{section:intro}.   
%should CWBG reasoning should be given for classes selection??

\subsubsection{Kinetics-600 Subset (KS)}
\label{SubSec:CurriculumDatasets}
Kinetics-600 contains $\approx$480,000 (mostly-adult) activity videos with an average duration of 10 seconds taken from YouTube. While the dataset contains 600 different activities with 600 videos per class on average, they vary from atomic actions like "Squat" to hierarchical activities like "Playing poker". Composition of the annotated kinetics-600-Subset is given in Table~\ref{T:child_adult_kinetics_600}. Four protocols were introduced for the model development and random 75\% train set, 25\% test set splitting was done.

\begin{itemize}
    \item KS-Full: Child data of all eight classes included. 
    \item KS-Large: Only the five classes with highest child data percentage are included (i.e. first five rows of Table~\ref{T:child_adult_kinetics_600}).
    \item KS-Balanced: Child data of same five classes as above but random sampling of 250 videos per class was done to create a balanced dataset.
    \item KS-Small-C: Only child data of three classes with lowest child data percentage are included (i.e. last three rows of Table~\ref{T:child_adult_kinetics_600}).
    \item KS-Small-A: Same three classes as in KS-Small-C, but contains only adult data.
\end{itemize}

\begin{table}[htbp] 
\begin{center}
\caption{Child-Adult video distribution of KS}
\label{T:child_adult_kinetics_600}
\resizebox{\columnwidth}{!}{%
\begin{tabular}{lccc}
\hline
\addlinespace[2pt]
\textbf{Class Name}             & \textbf{Child Data} & \textbf{Adult Data} & \textbf{Child Percentage} \\ \hline

Hopscotch              & 643        & 135        & 83\%             \\
Clapping               & 386        & 157        & 71\%             \\
Bouncing on trampoline & 534        & 315        & 63\%             \\
Baseball throw         & 293        & 256        & 53\%  
\\
Climbing tree          & 438        & 390        & 53\%             \\
Cutting watermelon     & 27         & 723        & 4\%              \\
Squat                  & 16         & 974        & 2\%              \\
Pull ups               & 59         & 704        & 8\%              \\
\addlinespace[1pt] 
\hline
\end{tabular}
}
\end{center}
\vspace{-3mm}
\end{table}

\subsubsection{Kinetics-Skeleton Subset (KSS)}
\label{SubSec:openpose_kinetics}
Kinetics-skeleton is a benchmark dataset used in HAR research that contains OpenPose-COCO extracted skeleton data of Kinetics-400 dataset. Due to the high computational cost associated with OpenPose extraction, this already extracted skeleton dataset was used in preliminary model building with Kinectics-400 and for the pre-training of ST-GCN model.
Since majority of the Kinetics-400 data are shared data with Kinetics-600, implementations on child data were also done according to the following protocols using the shared-annotated data.

% \begin{itemize}
%     \item KSS-Full: All shared-annotated child data of 8 classes included. 
%     \item KSS-Large: Only the shared-annotated child data of five classes as in KS-Large protocol.
%     \item KSS-Balanced: Shared-annotated child data of same five classes as above but random sampling of 110 videos per class was done to create a balanced dataset.
%     \item KSS-Small-C: This contains only the shared-annotated child data of three classes as in KS-Small-C.
%     \item KSS-Small-A: Same three classes as above but contains the shared-annotated adult data as well as the other unannotated data.
% \end{itemize}
We introduce KSS protocols with the same naming convention (e.g., KSS-Full for full 8 subset implementation of kinetics-skeleton child data). KSS-Balanced protocol contains 110 samples per each of the 5 class.

\subsection{CAR on Skeleton Modality}
\label{Sec:stgcn_main}

\begin{figure}[tbp]
    \centering
    \includegraphics[width=7cm]{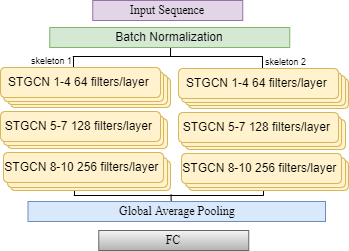}
    \caption{ST-GCN Model Architecture}
    \label{fig:stgcn_architecture}
    \vspace{-3mm}
\end{figure}

TensorFlow based ST-GCN model implementation from our   work~\cite{sanka2022icit} is taken as the base for this research. While the number of layers were kept same, number of filters was taken as a parameter that was determined in the first stage of the implementations. Feature vector in each node contains $x,y$ pixel coordinates and the confidence value resulted in OpenPose. Since we are considering multi-person scenarios in our model, we run two ST-GCN models parallely on each skeleton features but the convolution weights are shared among each model (Figure~\ref{fig:stgcn_architecture}). The Global Average Pooling layer stacked on top of GCN layers combine the outputs for two skeletons in each video resulting in a single vector which is used as the input to the fully connected (FC) classifier.

Though tracking of skeletons is important in multi-person scenarios, as the actions considered are mostly single-person, we used a basic tracking mechanism in calculating the minimum Euclidean distance of skeletons in this paper. %As tracking functionality is not available in OpenPose. tracking was done through calculating the euclidean distance of skeletons between consecutive frames and the minimum distance resulted for a particular skeleton was taken as the real distance between the particular skeleton in temporal domain. 
After normalizing the $x,y$ feature values, centralization was done with changing the origin of coordinate system to the center of image. Camera movement was simulated as detailed in the ST-GCN paper~\cite{yan2018spatial} through affine transformations like rotation, translation and scaling. For videos with less than 10 seconds duration, zero padding was implemented. %done to get the predefined 300 frames. 
Zeros were assigned for missing joints and sorting of skeleton sequence in a multi-person scenario was also implemented. 
\subsubsection{Learning Methods}
\label{SubSec:Learning}
Model training was first done without any pre-trained data as detailed in Section~\ref{subsubsection: vanilla_imp}. To improve the model performance, it was pre-trained with kinetics-skeleton dataset which can be considered as a special case of propagation approach detailed in~\cite{sanka2022icit}. Furthermore, both feature extraction (FX) and fine-tuning (FT) methods~\cite{sanka2022icit} were implemented using kinetics-skeleton as the source dataset. Applying the best performance FT method~\cite{sanka2022icit}, only the last GCN layer was kept trainable. 

%For the Fine-Tuning method, only the frozen layer approach was implemented since that resulted in best performance in~\cite{sanka2022icit}. With fine tuning, only the last GCN layer was kept trainable since detailed study on layer-wise fine-tuning done in~\cite{sanka2022icit} suggest that is the best configuration.   

\subsubsection{Skeleton Structures}
\label{SubSec:skeleton}
OpenPose provides two pose estimation models called BODY\_25 (i.e., OpenPose 2019 version) and COCO (i.e., OpenPose 2016 version). The BODY\_25 is %2X 
faster than COCO in extraction process and, its accuracy is also improved by 7\% as detailed in~\cite{cao2017realtime}. % compared to COCO.

COCO-skeleton ${(G_{coco})}$ contains 18 vertices$(V_{1})$ and BODY\_25-skeleton ${(G_{body})}$ contain 25 vertices$(V_{2})$ and since $V_{1}$$\subset$$V_{2}$ in ${G_{coco}}=(V_{1},E_{1})$ and ${G_{body}}=(V_{2},E_{2})$, ${G_{coco}}$ skeleton structure could be used with the BODY\_25 extracted data. Thus, skeleton extraction was done for the full kinetics600 subset using BODY\_25 model. 

\begin{figure}[tbp]
    \centering
    \includegraphics[width=8cm]{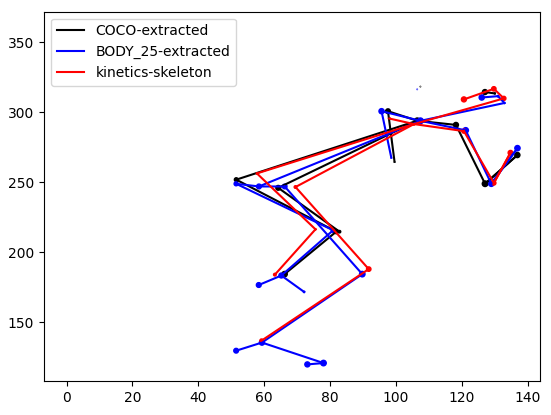}
    \caption{Skeleton structures from OpenPose}
    \label{fig:skeleton_compare}
    \vspace{-3mm}
\end{figure}

Since kinetics-skeleton is created with COCO model and graph structure has to be same in all data, we used ${G_{coco}}$ for KSS pre-trained model based implementations. Figure~\ref{fig:skeleton_compare} compares the skeleton structures for a frame taken from a "climbing tree" class video and the black colour represent the skeleton in kinetics-skeleton dataset, blue for the COCO  skeleton and red for the BODY\_25 skeleton. 

\subsubsection{Confidence Evaluations}
\label{SubSec:confidence}
For each joint $(J_{n})$, OpenPose outputs a 3 element vector containing $[x_{n},y_{n},C_{n}]$ where $C_{n}$ denotes confidence value and $C_{n}\subset[0,1]$  . Though $C_{n}$ can be used as a heuristic to analyse the effect of occlusion and truncation in videos %since such scenarios result in $C_{n}=0$. Yet any 
misidentified skeletons and unusual poses can't be analysed. %through this.
We calculate the average confidence of entire skeleton video for each person visible in the video. The calculation is further improved by removing skeleton-less frames and %Further in order to gain how visible joints affect results, we 
taking only the visible joints for the average. Extending these calculations to all sequences of a class, correlation of 'class-wise accuracy' and 'class-wise average confidence' variables were analysed using visualization, Pearson correlation and Spearman's correlation values. 

\subsection{CAR on RGB Modality}
%Action recognition process in the original LRCN architecture~\cite{donahue2015long} extracted visual features through CNN and learnt the sequential process through LSTM. Furthermore, pre-trained ImageNet model~\cite{deng2009imagenet}, which is commonly used in HAR with RGB videos~\cite{wang2019i3d},~\cite{ng2018actionflownet},~\cite{tran2018closer} has been used as the base model. %The pretrained ImageNet models were typically used .
%The LRCN base implementation was done on the benchmarked UCF101 dataset~\cite{donahue2015long}. The LRCN model used in this study was an improved LRCN approach that achieved the best results on action recognition using RGB videos.

Action recognition in the original LRCN architecture used CNN and LSTM and was pre-trained on a subset of ImageNet dataset~\cite{donahue2015long}. %has been used as the base model which improves the overall accuracy~\cite{donahue2015long}. The LRCN base implementation was done on the benchmarked UCF101 dataset~\cite{soomro2012ucf101}. The LRCN model used in this study was an improved LRCN approach that achieved the best results on action recognition using RGB videos~\cite{github_human_activity_recognition_LRCN_2020}.
The LRCN approach used in this study was based on~\cite{github_human_activity_recognition_LRCN_2020}. %which was trained on the UCF-101 dataset~\cite{soomro2012ucf101}. %the ImageNet dataset pre-trained on ResNet152~\cite{he2016deep} method, which provides strong initialization to encourage a faster training process and avoid overfitting on small video activity recognition. The input to the LRCN model was a preprocessed short video fragment representing the activity, and the output was the probability of possible vectors.The preprocessed video frames were embedded by the pre-trained model to extract visual features, and the sequence learning process was conducted by the bi-directional LSTM layer, output layer output the labels of the performed action~\cite{github_human_activity_recognition_LRCN_2020}.
It was initially trained and tested with the UCF101 dataset~\cite{soomro2012ucf101} for a selected subset of 55 classes. %to check the performance with the original LRCN paper implementation~\cite{donahue2015long}.Furthermore, the used model achieved considerable accuracy in that implementation, the same model was directly trained and tested with a selected subset of child data in the kinetics600 dataset~\cite{carreira2018short}.The LRCN model performed well in the main three implementations of child action recognition (8 class, 5 class, 3 class), the experiments focused on the direct approach.

\section{Experiments}
\label{section:experiments}

This section discusses the stages followed in conducting
the quantitative analysis including pre-processing of the datasets, and the experimental settings.

\subsection{Implementation Details}
\label{SubSec:__addNameLater__}
\subsubsection{ST-GCN}
\label{subsection: exp_st_gcn}
The ST-GCN architecture was kept the same as in~\cite{yan2018spatial} in all implementations since early experiments with number of filters and layers didn't result in consistent improvements. A 'piece-wise constant decay' function was used as the learning rate scheduler where the learning rate was reduced with a rate of 0.1 after a given number of steps, while the initial learning rate and the other hyper-parameters were varied in each implementation. In vanilla implementations with KS dataset, base learning rate was 0.001 and for transfer learning (TFL) implementations it was 0.1. Though the learning rate was same for TFL implementations, it was varied with each vanilla implementation for KSS. %But in KSS based vanilla implementations, this was different for each implementation but for transfer learning implementations it was 0.1. 
Stochastic Gradient Descent (SGD) as the optimizer,  categorical cross entropy as the loss function,  and a batch size of 4 were used with number of epochs being either 30 or 50 in all implementations. %in every implementation. Batch size of 4 was used and number of epochs were either 30 or 50 in all implementations.

\subsubsection{LRCN}
\label{subsection: exp_lrcn}
The LRCN implementation applied in this research used the same parameters tested for UCF101~\cite{github_human_activity_recognition_LRCN_2020}. 
In the data preprocessing stage, sampling rate was 10, FPS of videos were 25, and the number of frames extracted was 15 for preprocessing of each video. The main configurable parameters of the model architecture were, the learning rate initially set to $0.0005$, number of epochs $100$, and the batch size was $16$. %Furthermore, the above-mentioned parameters were used in all implementations and kept the same as in the ~\cite{github_human_activity_recognition_LRCN_2020}.

\subsection{Evaluation methods}
\label{subsection: evaluation_methods}
Top-1 accuracy as the main evaluation method together with confusion matrices were used to analyse the class-wise performance. Box-and-whisker plots were used to compare different learning methods in skeleton modality implementations and confidence interval calculations and visualization were done to compare different model performance.

\section{Results and Discussion} 
\label{section:results,discussion}
Initial implementations were done to determine the primary configurations of the ST-GCN model. Then the secondary configuration (i.e., graph structure, pre-processing etc.) were selected for best performance and the resultant configurations were used throughout the rest of the implementations. Parallelly, kinetics-skeleton implementation was done on ST-GCN to validate the model and to use as a pre-trained model.
%LRCN implementation using full kinetics-400/600 not possible due to computational cost - data ~500GB and train take ~40 days 

\subsection{Skeleton Modality Results}
\label{subsection: skeleton_modality}
%intro about skeleton modality
\subsubsection{Preliminary Implementations}
\label{subsection: preliminary_imp}
Since all the 8 interested activities are supposed to be done by a single person, performance was compared between one-person implementation (1-person Model) and two-people implementation (Standard Model) on KSS Full dataset. Implementation was further extended by discarding the confidence value (Section~\ref{SubSec:confidence}) from joint feature vectors  (2D only Model). %These were all done on the using all data of 8 classes subset of kinetics-skeleton dataset.
Since the standard model outperforms the 1-person model (Table~\ref{T:configuration}), 2 people per frame was chosen for number of people per frame configuration value. Though disregarding confidence feature values improves the accuracy (Table~\ref{T:configuration}), 2D+confidence (i.e., standard model) feature vector was chosen as the default configuration since the improvement is marginal and we need to compare different key point extraction based models.   
% we use kinetics-skeleton since STGCN authers have released the pre-processed data as well so issue with pre-process

\begin{table}[htbp] %h- here , t-top , b- bottom, p- page (page dedicated to figures, tables etc.)
\begin{center}
\caption{Preliminary Implementation}
\label{T:configuration}
\begin{tabular}{lcc} %left aligned the first col and center the second
\hline
\textbf{Implementation} & \textbf{Top-1 Accuracy} & \textbf{Top-5 Accuracy}\\
\hline
\addlinespace[2pt]
Standard Model          & 66.61          & -              \\
1-person Model          & 63.29          & -              \\
2D only Model           & 68.34          & -              \\ \hline
Kinetics-skeleton [original] & 30.7           & 52.8           \\
Kinetics-skeleton            & 21.16          & 41.7           \\ 
\addlinespace[1pt] 
\hline
\end{tabular}
\end{center}
\vspace{-3mm}
\end{table}
%---kinetics400 full
Full implementation of kinetics-skeleton was done but the data of the 8 classes were not used. Since that only account for 0.6\% of kinetics-skeleton dataset, effect of this on model performance should be minimum. Original hyperparameters were used where it was possible but some (e.g., batch size) were not used due to computational limitations. This or the differences in pre-processing may have lead to the low performance in Table~\ref{T:configuration}.  
%---kinetics400 full

For further comparison, implementations were done on the skeleton structure and between adult and child subsets as well. For these implementations, a balanced 4 class subsets were used with 98 samples per class. A  modified graph structure instead of OpenPose-COCO graph structure where hip joints are connected to corresponding shoulder joints rather than the neck joint (Table~\ref{T:secondary_configuration}-Modified structure) was experimented which resulted in better performance. 
%Based on the results in Table~\ref{T:secondary_configuration}, we decided to use a modified graph structure instead of OpenPose-COCO graph structure where hip joints are connected to corresponding shoulder joints rather than the neck joint. 
Thus, we attempted this configuration with the kinetics-skeleton pre-trained model which resulted in a sharp improvement of accuracy, both in child and adult dataset implementations (Table~\ref{T:secondary_configuration}-Modified pre-trained). Hence later implementations were attempted with kinetics-skeleton pre-trained implementations as well.% may be we can compare child vs adult results to argue that the bottleneck happens not due to child vs adult thing (ex: pose estimation error due to anatomical diffferences) but rather the overall skeleton extraction is not good due to occlusion , tracking issue etc.

\begin{table}[htbp] 
\begin{center}
\caption{Secondary Implementation}
\label{T:secondary_configuration}
\begin{tabular}{lcc} %left aligned the first col and center the second
\hline
\addlinespace[2pt]
\textbf{Implementation}       & \textbf{Child Dataset} & \textbf{Adult Dataset} \\ \hline
OpenPose-COCO structure   &  63.26 & 46.93 \\
\textbf{Modified structure}   & 61.22  & 56.12   \\
Modified pre-trained & 86.73 & 84.69 \\ \hline
Random frame selection [$w=150$]      & 55.10 & 43.87     \\
Random skeleton movement           & 61.12 & 61.22     \\
\textbf{Combined approach} [$w=150$]      & 62.24 & 58.16 \\ 
Sub-sampling approach & 50.0  & 44.89 \\ 
\addlinespace[1pt] 
\hline
\end{tabular}
\end{center}
\vspace{-3mm}
\end{table}

Experiments with final pre-processing stage were also done to select the best configuration. Authors of original ST-GCN paper~\cite{yan2018spatial} had implemented random frame selection process as well as random skeleton movement process, thus we first implemented them separately. Random frame selection was done with a window size ($w$) of 150 and gradually increased to 250, but the best results we achieved ($w=150$) were below the performance of "Modified structure" approach (Table~\ref{T:secondary_configuration}). As the best performance was achieved when both processes were combined   (Table~\ref{T:secondary_configuration}-Combined approach), it was considered as the default configuration. In addition, 
%But when these were combined, we were able to achieve better performance thus the "combined approach" was selected as the configuration. 
a new sub-sampling approach with frame dropping was also implemented, but due to low performance it was not added to the final pre-process stage.
 
For KS based implementations, comparisons were done between different skeleton structures (Table~\ref{T:skeleton_K600}). Feet related joints were removed from $G_{body}$ resulting in $G_{body}\ast$ skeleton structure with 19 vertices $(V_{3})$. Considering the overall performance, $G_{body}$ was used for the vanilla implementations while $G_{coco}$ was used with pre-trained model implementations. 

\begin{table}[htbp] 
\begin{center}
\caption{Skeleton Structure Selection}
\label{T:skeleton_K600}
\begin{tabular}{lcc} 
\hline
\addlinespace[2pt]
\textbf{Skeleton structure}       & \textbf{KS-balanced} & \textbf{KS-Full} \\ \hline
$G_{body}$ , $|V_{2}|=25$        & 69\%          & 75\%      \\
$G_{body}\ast$ , $|V_{3}|=19$                  & 67\%          & 74\%      \\
$G_{coco}$ , $|V_{1}|=18$                 & 66.45\%         & 74\%      \\ 
\addlinespace[1pt] 
\hline
\end{tabular}
\end{center}
\vspace{-3mm}
\end{table}

\subsubsection{Vanilla implementation on Child data}
\label{subsubsection: vanilla_imp}
Implementations were done on both KS and KSS protocols and the models were not pre-trained on any other dataset. %None of the transfer learning techniques were utilized in these implementations. Batch size of 4 was used due to memory limitations and other hyper-parameter tuning was done.
Results in Table~\ref{T:vanilla_stgnc} suggest a general improvement in accuracy when moving from a KSS protocol to a KS protocol. This improvement could be due to the higher number of data samples per class in KS or due to the improvement of graph structure as demonstrated by the results in Table~\ref{T:skeleton_K600}. When comparing the class-wise accuracy between KS/KSS-Balanced protocols, performance of each class has increased, yet there is no relative improvement between classes.'Clapping' class perform the best while 'baseball throw' perform the worst. Result also suggest there is no strong connection between confidence value and accuracy given that 'climbing tree' class performs second best even though the average confidence value is the lowest and 'hopscotch' performs second worst even though the average confidence is the highest. 

\begin{table}[htbp] 
\begin{center}
\caption{Vanilla Implementation Results}
\label{T:vanilla_stgnc}
\begin{tabular}{llllll}
\hline
\addlinespace[2pt]
\textbf{From Scratch} & \textbf{Full}  & \textbf{Large} & \textbf{Balanced} & \textbf{Small-C} & \textbf{Small-A} \\
\hline
KS                   & 75.29 & 77.83 & 69.32    & 69.23 & 92.34 \\
KSS                   & 60.68 & 64.88 & 59.43    & 86.95 & -   \\
\addlinespace[1pt] 
\hline
\end{tabular}
\end{center}
\vspace{-3mm}
\end{table}

\subsubsection{Kinetics-Skeleton Implementation}
\label{subsubsection: kinetics-skeleton_imp}
Class-wise evaluation of the kinetics-skeleton implementation introduced in Section~\ref{subsection: preliminary_imp} was done and the results are given in the Table~\ref{T:kinetics_skeleton_analysis} for the 8-class subset. Class index refers to the class index of the Table~\ref{T:child_adult_kinetics_600}. `Position' refers to the place each class take when all 400 classes are ordered in descending order in terms of class-wise accuracy. Confidence value is calculated per person in video and since there are only 2 people maximum, both of these values are given in this row.  

\begin{figure}[tbp]
    \centering
    \includegraphics[width=8cm]{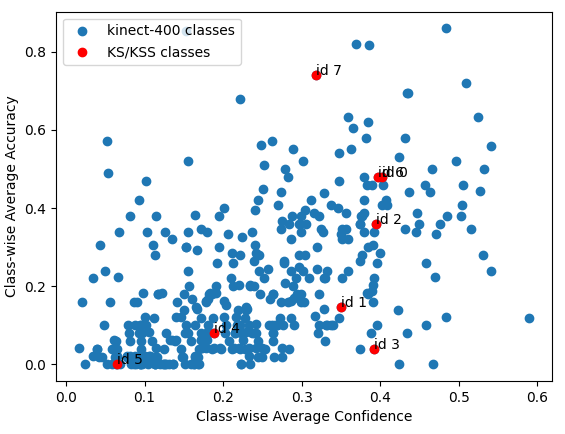}
    \caption{Accuracy vs Confidence Comparison}
    \label{fig:acc_vs_conf}
    \vspace{-3mm}
\end{figure}

Higher accuracy and position attained by classes with indices 7, 6, 0, and 2 (Table~\ref{T:child_adult_kinetics_600}) can be explained as a result of motion-oriented nature of those actions. Considering the distribution of all Kinetics-skeleton 400 classes in accuracy vs confidence (Figure~\ref{fig:acc_vs_conf}), all four classes are above average. Relatively bad performance of other 4 classes is difficult to attribute to a single cause. Considering 4 and 5 classes, it may be due to truncation/occlusion present in the videos as evidenced by the confidence values but same reasoning %suggest it is not the cause 
is not true for the low performance of 1 and 3 classes.    

Visualized distribution of all 400 classes in Figure~\ref{fig:acc_vs_conf} implies there is no strong correlation but if the average confidence values is close to zero, then there is a higher chance of resulting in a low accuracy. Quantitative analysis resulted in 0.533 for Pearson correlation indicating only a moderate positive relationship and 0.564 for Spearman's correlation. Analysis on other implementation also resulted in similar results and conclusions. 
%While clapping and baseball throw are also motion oriented, it involves less amount of body part movement and consequently the videos may contain higher amount of truncation. This can be explained through the average confidence value resulted for these two classes. 

\begin{table}[htbp] 
\begin{center}
\caption{kinetics-skeleton results for 8 classes}
\label{T:kinetics_skeleton_analysis}
\resizebox{\columnwidth}{!}{%
\begin{tabular}{lllllllll}
\hline
\addlinespace[2pt]
\textbf{Class Index} & \textbf{0}  & \textbf{1} & \textbf{2} & \textbf{3} & \textbf{4} & \textbf{5} & \textbf{6} & \textbf{7} \\ 
\hline
% Class      & 0       & 1       & 2       & 3       & 4       & 5       & 6       & 7       \\
Accuracy   & 48\%    & 14.58\% & 36\%    & 4\%     & 8\%     & 0.00\%  & 48\%    & 74\%    \\
Position      & 32(2)   & 210(5)  & 79(4)   & 321(7)  & 276(6)  & 373(8)  & 34(3)   & 5(1)    \\
Confidence & .40/.12 & .35/.18 & .39/.11 & .39/.16 & .19/.03 & .06/.01 & .40/.12 & .32/.05\\
%Place                   & 60.68 & 64.88 & 59.43    & 86.95 & -  & 69.32    & 69.23 & 92.34\\
\addlinespace[1pt] 
\hline
\end{tabular}%
}
\end{center}
\vspace{-3mm}
\end{table}

\subsubsection{Transfer Learning Implementations}
\label{subsubsection: K400-X_imp}
% contain all the implementation results discussion of TFL/pre-trained implementations

Initial implementations with propagation approach resulted in considerable accuracy improvement for both KS and KSS implementations. This can be attributed to the Kinetics-skeleton dataset size and diversity. FX and FT approaches using Kinetics-skeleton as the source dataset resulted in marginal differences compared to propagation approach. But in our previous work~\cite{sanka2022icit}, both FX and FT outperformed propagation approach and this could be either due to increased diversity in source dataset in this implementation compared to the NTU dataset used as the source dataset in~\cite{sanka2022icit} or simply due to the presence of negative TFL in these implementations. Further experiments are needed to validate these claims. 

To analyse these approaches in detail, KSS-Full protocol class-wise results for each approach was visualized using a box-and-whisker plot in Figure~\ref{fig:class_probability_compare}. Resultant probability value from ST-GCN model's final softmax-activation based dense layer was used as the sample probability to develop the probability distribution of each class. Transfer Learning approaches class-wise gain over vanilla approach is evident through the median values of box plot. Based on the median values, propagation approach performs better than other TFL methods even though the difference is small. When comparing the distribution variance of each approach, propagation approach has a considerably low variance in `pull ups' class and the large five classes with the  exception of `baseball throw' class. Since all these distributions can be interpreted as combinations of correctly classified sample distribution and misclassified sample distribution, the high variance of `throw baseball' can be explained as a result of considerable contribution of misclassified sample distribution. While the class-wise accuracy improvement of 5 and 6 classes are negligible, with propagation approach, upper quartile has increased considerably, implying there is a room for improvement if more data is present.

\begin{table}[tbp] 
\begin{center}
\caption{Transfer Learning Results}
\label{T:tfl_results}
\resizebox{\columnwidth}{!}{%
\begin{tabular}{llcccc} 
\hline
\addlinespace[2pt]
&\textbf{Implementation}     & \textbf{Full}  & \textbf{Balanced} & \textbf{Large} & \textbf{Small-C} \\ \hline
{\multirow{1}{*}{\rotatebox[origin=c]{0}{KS}}}&Propagation        & 84.3 & 83.38 & 86.03    & 76.92 \\
%&Fine-Tuning        & 80.47 & 89.85 & 86.51    & 82.60 \\
%&Feature Extraction & 79.15 & 89.13 & 87.92    & 86.95 \\
\hline
{\multirow{3}{*}{\rotatebox[origin=c]{0}{KSS}}}&Propagation        & 81.26 & 87.68 & 87.92    & 82.60 \\
&Fine-Tuning        & 80.47 & 89.85 & 86.51    & 82.60 \\
&Feature Extraction & 79.15 & 89.13 & 87.92    & 86.95 \\
\hline
\addlinespace[1pt] 
\end{tabular}%
}
\end{center}
\vspace{-5mm}
\end{table}

\begin{figure}[htbp]
    \centering
    \captionsetup{justification=centering}
    \includegraphics[width=9cm]{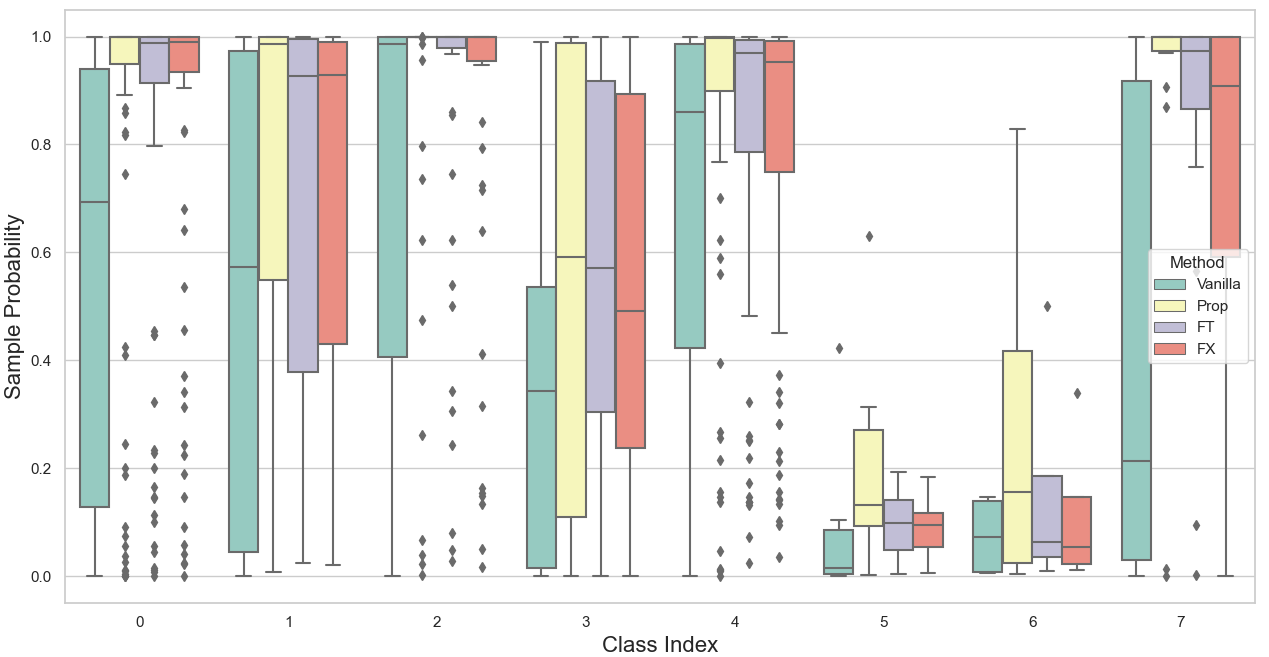}
    \caption{Probability Distribution of Skeleton Modality Implementations}
    \label{fig:class_probability_compare_8}
    \vspace{-3mm}
\end{figure}

\subsection{RGB Modality Results}
\label{subsection: RGB_modality}

RGB implementations were carried out on the same dataset protocols as given in Section~\ref{section:Datasets}. The performance comparison between different datasets shows a good accuracy in RGB implementation except for KS-Small-C dataset due to low number of samples( Table~\ref{T:RGB_SkeletonModels}). When compared to the other datasets, the balanced dataset achieves the best accuracy for both KSS and KS protocol implementations. Furthermore, the use of pre-trained ResNet-152 in LRCN enhanced the model's training efficiency and prevented overfitting for lower number of samples.

\begin{table}[htbp]
\begin{center}
\captionsetup{justification=centering}
\caption{Comparison of RGB modality and Skeleton modality}
\label{T:RGB_SkeletonModels}
\begin{tabular}{lccccc}
\hline
\multicolumn{1}{l}{\textbf{Dataset}} & \multicolumn{1}{c}{\textbf{Modality}} & \multicolumn{1}{l}{\textbf{Full}} & \multicolumn{1}{l}{\textbf{Balanced}} &
\multicolumn{1}{l}{\textbf{Large}} &
\multicolumn{1}{l}{\textbf{Small-C}} \\ \hline
{\multirow{2}{*}{\rotatebox[origin=c]{0}{KS}}}      & RGB           & 86.62                             & 88.64        & 87.02                         & 73.07                                \\ 
 & Skeleton           & 84.3                       & 83.38      & 86.03                           & 76.92                                \\
 \hline
 \addlinespace[2pt]
{\multirow{2}{*}{\rotatebox[origin=c]{0}{KSS}}}    & RGB                 & 82.57                             & 86.23               & 79.72                  & 78.26                                \\
& Skeleton                 & 81.26               & 89.85        & 87.92                         & 86.95                                \\
\hline
\end{tabular}
\end{center}
\vspace{-5mm}
\end{table}

\subsection{Model Comparison}
\label{subsection: modal_comp}
Since LRCN model is pre-trained on the ImageNet dataset, comparison of ST-GCN vanilla implementation with LRCN is not sensible. Instead we compare the pre-trained ST-GCN model based implementations (i.e., propagation approach) with the pre-trained LRCN implementation. Between the KS\slash KSS protocol based implementations, the differences are marginal as detailed in Table~\ref{T:RGB_SkeletonModels}. While the LRCN performs better than ST-GCN in KS-Balanced protocol, the opposite is true regarding the KS-Small protocol. Comparing the KSS protocols, ST-GCN performs better than LRCN in both KSS-Balanced and KSS-Small-C protocols. Since the differences are marginal, we argue that performance of both models are comparable.   

\begin{figure}[htbp]
    \centering
    \captionsetup{justification=centering}
    \includegraphics[width=8cm]{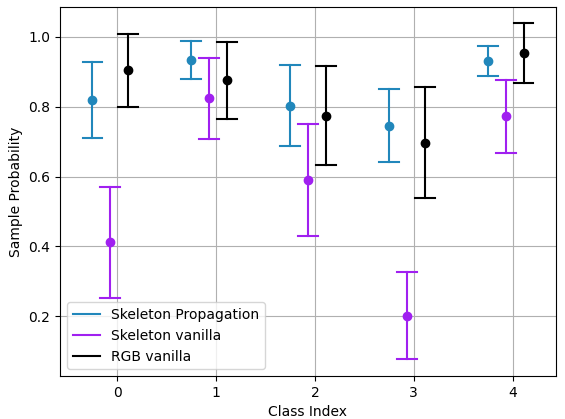}
    \caption{Performance of Models with Confidence Interval}
    \label{fig:class_probability_compare}
    \vspace{-4mm}
\end{figure}

A class-wise sample probability comparison was done with confidence interval (CI) of 95\% for KSS-Balanced protocol. Result in Figure~\ref{fig:class_probability_compare} suggest that LRCN performs better in `hopscotch' and `climbing tree' classes but ST-GCN performs better in all others. Thus, even though in terms of Top-1 accuracy , ST-GCN performs better, these results also show that overall performance is similar. In some instances upper limit of CI goes beyond 1 as a result of Gaussian assumption in CI calculation. 

In order to compare 2D Skeleton and RGB modality results with similar classes of 3D data, for KS/KSS-Full protocol, 8 similar class implementations using the CWBG dataset were done with the ST-GCN model used in~\cite{sanka2022icit} and the best result was achieved with FT approach. The 3D result is shown as `Kinect-Camera' in Table~\ref{T:model_acc_compare} along with best results we achieved in this work.

\begin{table}[tbp] 
\begin{center}
\caption{Comparison of Models}
\label{T:model_acc_compare}
\begin{tabular}{lccc}
\addlinespace[2pt]
\hline
\textbf{Model}    & \textbf{Kinect Camera} & \textbf{OpenPose Model} & \textbf{RGB Model} \\ \hline
Accuracy & 70.56      & 84.3           & 86.62     \\ \hline
\addlinespace[1pt] 
\end{tabular}
\end{center}
\vspace{-5mm}
\end{table}

\section{Conclusion and Future Works}
\label{section:conclusion}
Inspired by the superior performance of skeleton based models in HAR research, we carry out an in-depth analysis of CAR on pose-based GCN. This analysis is further extended with a  comparison of RGB modality and Skeleton modality in CAR. Though the average class-wise confidence values show that unconstrained nature of videos severely limit the pose estimation process, in this research we show that ST-GCN model is able to achieve comparable performance to LRCN model. %with the use of large dataset based pre-training. 

Based on our results on KS/KSS-Full protocol, LRCN performance over ST-GCN suggest a limitation of GCN when truncation and occlusion are present in videos. ST-GCN performance over LRCN on KS/KSS-Small-C protocol suggests a better discrimination ability of GCN when presented with small number of classes. On KS/KSS-Balanced protocol both models perform equally well suggesting  similar potential of LRCN and ST-GCN model architectures.
% On KS/KSS-Full protocol, LRCN achieved the best performance of 86.62\%. ST-GCN achieved 89.85\% on KS/KSS-Balanced protocol and 86.95\% on the three class KS/KSS-Small-C protocol as the best performance. 

While the average confidence value intuitively quantify the `visibility' of the skeleton sequence, Pearson correlation coefficient value of 0.53  between class-wise accuracy and average confidence value suggest that in addition to the average confidence value, presence of other latent variables affect skeleton modality performance.

%correlation is not strong enough thus implying the presence of other latent variable at play in skeleton modality performance. 

%final line
Despite the pose differences in adults and children, these results suggest that when the actions are motion oriented, skeleton modality can perform on par with RGB modality, thus opening future research directions in optimal skeleton-graph creation, pose estimation improvements in complex scenarios and developing ensemble of multiple modalities. 
% and skeleton-graph interpretation methods.           

%LRCN model was used with RGB modality while ST-GCN with skeleton modality and experiments were conducted on a manually annotated Kinect-600 based dataset. Comparison of these two modalities were done using four protocols.

% While the RGB approach (pre-trained with ImageNet) outperformed the vanilla implementation of skeleton modality, when skeleton modality was   

%no 
%%%%%%%%%%%%%%%%%%%%%%%%%%%%%%%%%%%%%%%%%%%%%%%%%%%%%%%%%%%%%%%%%%%%%%%%%%%%%%%%
% \newpage
\small{   
\section*{Acknowledgment}
This research was supported by the Accelerating Higher Education Expansion and Development (AHEAD) Operation of the Ministry of Higher Education of Sri Lanka funded by the World Bank (\url{https://ahead.lk/result-area-3/}).
}
%\bibliographystyle{ieee}
% \bibliography{egbib}
\bibliographystyle{IEEEtran}
\bibliography{references}

\end{document}